\documentclass[11pt]{article}
\usepackage[preprint]{acl}
\usepackage{times}
\usepackage{latexsym}
\usepackage[T1]{fontenc}
\usepackage[utf8]{inputenc}
\usepackage{microtype}
\usepackage{inconsolata}
\usepackage{graphicx}
\usepackage{booktabs}
\usepackage{array}
\usepackage{enumitem}
\usepackage{url}
\usepackage{xurl}
\usepackage{amsmath}

\title{What Do Rationales Communicate? A Message-Intervention Study in Role-Specialized QA}
\author{
Jiameng Zhang \\
University of Zurich \\
\texttt{jiameng.zhang@uzh.ch}
\And
Hongqiu Wu \\
Shanghai Jiao Tong University \\
\texttt{wuhongqiu@sjtu.edu.cn}
}

\begin{document}
\maketitle

\begin{abstract}
Role-specialized QA pipelines increasingly pass rationales from a reasoner to a verifier, but it is unclear what this message actually buys: better answers, stronger support assessment, or a new failure surface. We introduce a message-intervention diagnostic that fixes the evidence and candidate answer while varying only the rationale passed across the reasoner-to-verifier boundary. On 400 MuSiQue, HotpotQA, and 2WikiMultiHopQA examples with DeepSeek as generator and verifier, faithful rationales add almost no answer accuracy over no rationale, while corrupted rationales strongly alter support judgments. Under a blind verifier prompt, harmless paraphrases shift support by only 0--2.5\%, whereas corrupted rationales shift support by 10--22\%; an explicit rationale-checking prompt amplifies the same pattern to 34--55\%. Final answers move less (2--30\%), and only 2.9--35.3\% of corrupted support flips co-occur with answer changes. Human audits show why this matters: 16/42 valid corruptions are corruption-overtrust cases, and blind humans reject or mark unclear 9/10 audited corrupted rationales that the model accepts. Cross-model and task-boundary checks show when the channel is active, amplified, inert, or folded into the task label. Rationale sharing should be evaluated as a verification-message mechanism, not merely as a route to higher answer accuracy.
\end{abstract}

\section{Introduction}

Many deployed LLM question-answering systems are role-specialized pipelines: a retriever selects evidence, a reasoner proposes an answer and a rationale, and a verifier judges whether the answer is supported. These roles may run on the same underlying model, but they communicate across explicit call boundaries with distinct prompts, inputs, and outputs, as in recent agent frameworks and reasoning-and-acting pipelines \citep{wu2023autogen,li2023camel,yao2023react,shinn2023reflexion,tran2025multiagent}. The usual reason for passing a rationale is simple: it should help the verifier use the evidence better and reach a better final answer.

That design choice hides an unresolved interface question: once an answer has already been proposed, what does the rationale message actually change? If it improves answer selection, then unsupported outputs should trigger answer regeneration. If it mainly changes support assessment, then the rationale is better treated as a claim to verify, repair, or ignore. Existing end-task accuracy cannot distinguish these cases because a verifier usually returns two things, not one: an answer and a judgment that the answer is supported.

\begin{figure}[t]
\centering
\includegraphics[width=0.96\columnwidth]{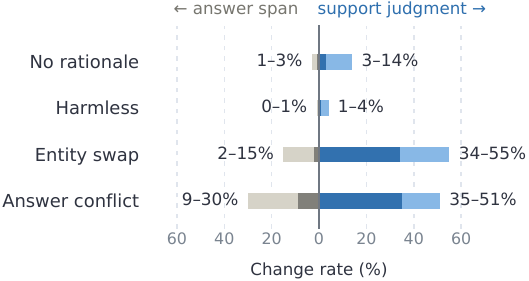}
\caption{Divergent sensitivity under a fixed-evidence, fixed-answer intervention. Only the rationale changes. Bars show min--max ranges across MuSiQue, HotpotQA, and 2Wiki: corrupted rationales move support much more than answers, while harmless paraphrases stay near zero.}
\label{fig:support-answer}
\end{figure}

We answer this question with a message-intervention protocol, summarized in Figure~\ref{fig:support-answer}. We treat the rationale as an explicit message sent from the reasoner to the verifier, not as a hidden trace of the model's thinking. Traditional chain-of-thought faithfulness work asks whether an explanation reflects the computation of the same model that produced an answer. We instead ask what happens when explanation-like text becomes a communication payload consumed by a later role. The protocol fixes the evidence and candidate answer, varies only the rationale, and measures answer selection and support assessment separately. Each example is verified under five conditions: no rationale, the original reasoner-generated rationale, a harmless meaning-preserving paraphrase, an entity-swapped corruption, and an answer-conflicting corruption.

Across MuSiQue, HotpotQA, and 2WikiMultiHopQA, the diagnostic shows that rationales are strongest as verification messages, not answer-selection signals. On MuSiQue, original rationales add only +0.5 EM over passing evidence and answer alone. Yet factual errors in the rationale sharply alter the verifier's support judgment. Under a blind verifier prompt, harmless paraphrases change support in only 0--2.5\% of examples, while corrupted rationales still change support in 10--22\%. When the system explicitly asks the verifier to check rationale faithfulness, the same channel becomes much stronger: corrupted rationales change support in 34--55\% of examples ($p<10^{-9}$ on every dataset). This is the setting where our diagnostic is most useful: pipelines that intentionally pass rationales for verification need to know whether that message field is active, harmless, or ignored.

The result gives a concrete design advantage. It separates three receiver states that final EM cannot see. An \emph{active} rationale channel changes support assessment and should be audited for overtrust. A \emph{harmless} channel leaves support stable under meaning-preserving rewording. An \emph{inert} channel, as in our DeepSeek-R1 boundary check, discounts the external rationale and may add interface complexity without communication value. This turns the vague design question ``should we pass rationales?'' into the measurable question ``what behavior does this message field change?''

We identify two auditable failure modes that final EM alone cannot see. The sharper one is \emph{corruption overtrust}: the verifier keeps accepting a corrupted rationale, and blind human labels reject or mark unclear 9/10 audited corrupted-supported cases. We also find \emph{correct-answer penalty}, where a correct answer is rejected because its rationale is corrupted. A third qualitative pattern, \emph{local answer dominance}, occurs when direct evidence preserves the answer despite a broken upstream rationale.

Our contribution is a diagnostic framework for role-specialized QA communication. We make three contributions:
\begin{enumerate}[leftmargin=*]
\item We introduce a message-intervention protocol for testing what a rationale changes after it is passed to a verifier.
\item We define paired metrics that separate answer selection from support assessment while holding evidence and candidate answers fixed.
\item We show across three multi-hop QA datasets that harmless paraphrases rarely change verifier judgments, while corrupted rationales substantially change support judgments; human audits, cross-model checks, blind-prompt checks, a repeated-run noise floor, and a claim-verification extension identify when this channel is active, amplified, inert, or task-coupled.
\end{enumerate}

Together, these contributions turn a vague design question, ``should a pipeline pass rationales to later roles?'', into a measurable one: what behavior does the rationale message change?

\section{Diagnostic Framework}

\subsection{Role-Specialized QA Pipeline}

We consider a three-role QA pipeline. A retriever selects evidence passages, a reasoner produces a candidate answer and explicit rationale, and a verifier receives the evidence, candidate answer, and optional rationale before returning a final answer and a binary support judgment. The roles may use the same underlying LLM family or different model families. Our object of study is the interface between calls: the message that is passed and the effect it has later. We treat rationales as explicit messages between roles, not as faithful traces of hidden model reasoning.

The diagnostic is a causal intervention on the message between roles. The task, evidence, candidate answer, verifier role, and output schema are fixed; only the rationale changes. If verifier behavior changes under this intervention, the change is attributable to the rationale field rather than to a different retriever, reasoner, or end-to-end compute budget.

\subsection{Perturbation Conditions}

Each example is evaluated under five conditions. In the \textbf{no-rationale} condition, the verifier receives only evidence and the candidate answer. In the \textbf{original-rationale} condition, it receives the reasoner-generated rationale. \textbf{Harmless paraphrase} rewords the rationale while preserving entities, relations, dates, and answer support. \textbf{Entity-swapped corruption} replaces a key entity, date, place, title, or relation with an incompatible alternative. \textbf{Answer-conflicting corruption} alters the rationale to imply a different plausible answer than the candidate answer.

These conditions separate two questions. Harmless paraphrases test whether the verifier is brittle to wording. Corrupted rationales test whether it responds to factual errors in the rationale.

\subsection{Metrics and Testing}

Let $r$ denote the original rationale condition and $c$ another condition. For example $i$, the verifier returns answer $A_i(c)$ and support judgment $S_i(c)$. The support judgment is a verifier output, not a proof-level entailment label. We use it to study communication behavior: whether a message changes what the verifier accepts as supported under a fixed prompt and evidence context. We report answer accuracy, support rate, paired sensitivities, and a coupling probability:
\[
\begin{aligned}
\mathrm{AnsSens}(c) &= \frac{1}{n}\sum_i \mathbf{1}[A_i(c)\neq A_i(r)],\\
\mathrm{SuppSens}(c) &= \frac{1}{n}\sum_i \mathbf{1}[S_i(c)\neq S_i(r)],\\
\mathrm{Couple}(c) &= P(\Delta A_i(c)\mid \Delta S_i(c)).
\end{aligned}
\]
Here $\mathrm{Couple}(c)$ asks how often answer changes accompany support flips. We avoid making the difference between answer-span changes and binary support changes carry the main argument, since the two outputs have different base rates and degrees of freedom. Harmless robustness means low support sensitivity under harmless paraphrase. Corruption responsiveness means high support sensitivity under factual corruption.

Because support judgments are paired binary outcomes, we use paired McNemar tests when comparing original rationales with harmless or corrupted conditions.

\section{Experimental Setup}

\subsection{Datasets and Models}

We evaluate on three multi-hop QA datasets: MuSiQue \citep{trivedi2022musique}, with 200 validation examples including 104 2-hop, 63 3-hop, and 33 4-hop questions; HotpotQA \citep{yang2018hotpotqa}, with 100 distractor-setting examples balanced between bridge and comparison questions; and 2WikiMultiHopQA \citep{ho2020constructing}, with 100 examples balanced across comparison, bridge-comparison, compositional, and inference categories. We additionally run a smaller SciFact claim-verification extension \citep{wadden2020scifact} as a boundary check. Unlike extractive QA, SciFact's output is itself a support/refute label, so it tests whether answer/support dissociation changes when the task output is tightly coupled to verification.

The main pipeline uses DeepSeek as generator/reasoner and verifier. Concretely, experiments were run through API endpoints in June--July 2026 with temperature 0: the main model identifier is \texttt{deepseek-v4-flash}; the Qwen robustness checks use \texttt{qwen3.7-plus-2026-05-26}; the Claude recheck uses \texttt{claude-haiku-4-5}; and the reasoning-verifier boundary check uses \texttt{deepseek-reasoner} (the DeepSeek-R1-family API identifier). Provider-side parameter counts and exact weight snapshots are not disclosed for these hosted models; Appendix~\ref{sec:model-details} lists the model strings and decoding settings used. We reuse existing full-history pipeline outputs for the candidate answer and original rationale. The verifier prompt asks the model to check whether the evidence supports the candidate answer, attend to rationale faithfulness if a rationale is supplied, and return JSON with final answer, support judgment, confidence, and note. To test prompt and model dependence, we run HotpotQA checks with Qwen as verifier, Claude as verifier, Qwen as generator/reasoner, and a DeepSeek-R1 verifier; we also run a blind verifier prompt on all three QA datasets. Hosted APIs do not expose seed control, so we estimate a test--retest noise floor by rerunning the faithful condition once on all 400 QA examples.

For the verifier-note analysis, we use a deterministic string check rather than an additional model judge: a note is counted as target-aware if it explicitly mentions the corrupted entity, date, or answer string, and as conflict-aware if it contains broader terms such as conflict, contradiction, inconsistency, incorrect, or mismatch. Target-aware matches are the stricter evidence that the verifier localizes the introduced semantic conflict; conflict-aware matches are a broader sanity check.

\subsection{Perturbation and Human Audits}

For each example, an LLM generates one harmless paraphrase and two corrupted rationales from the original rationale. These are LLM-assisted intervention stimuli, not automatically trusted labels; the design depends on their quality, so we audit them explicitly. A 48-item corrupted-rationale audit balanced by hop count and corruption type finds that 42/48 (87.5\%) are valid perturbations. A 50-item harmless-paraphrase audit finds that 36/50 (72.0\%) are strictly valid and 43/50 (86.0\%) are acceptable under a lenient criterion. We also run a 50-item human support audit enriched for diagnostic cases. A second annotator performs a blind pass, seeing the question, evidence, candidate/final answer, and displayed rationale, but not the model support label, verifier note, or sampling bucket. We use this blind pass as the primary human support label and report agreement by sampling bucket in Section~5.1; pooled model-human agreement is only descriptive because the audit is intentionally enriched. The two human annotation passes agree on 88\% of items ($\kappa=0.77$). This audit validates targeted support changes, not global verifier accuracy. Finally, to check that corruption is not merely a synthetic stress case, we audit 100 unperturbed reasoner rationales and find 11 unfaithful rationale errors plus 16 metric false negatives, showing that natural pipeline outputs already contain the kind of rationale/evidence mismatch our intervention isolates.

\section{Rationales Verify More Than They Answer}

\begin{table*}[t]
\centering
\scriptsize
\setlength{\tabcolsep}{3.2pt}
\begin{tabular}{@{}llrrrrr@{}}
\toprule
Dataset & Condition & EM & Support & $\Delta$Ans & $\Delta$Supp & $P(\Delta A\mid\Delta S)$ \\
\midrule
MuSiQue & No rationale & 51.5 & 86.5 & 2.0 & 14.0 & 7.1 \\
MuSiQue & Faithful & 52.0 & 85.5 & -- & -- & -- \\
MuSiQue & Harmless & 51.5 & 84.5 & 1.0 & 4.0 & 12.5 \\
MuSiQue & Entity swap & 48.0 & 46.5 & 6.5 & 41.0 & 8.5 \\
MuSiQue & Answer conflict & 48.5 & 45.0 & 12.0 & 42.5 & 15.3 \\
\midrule
HotpotQA & No rationale & 69.0 & 97.0 & 1.0 & 3.0 & 0.0 \\
HotpotQA & Faithful & 68.0 & 100.0 & -- & -- & -- \\
HotpotQA & Harmless & 68.0 & 99.0 & 0.0 & 1.0 & 0.0 \\
HotpotQA & Entity swap & 68.0 & 66.0 & 2.0 & 34.0 & 2.9 \\
HotpotQA & Answer conflict & 62.0 & 65.0 & 9.0 & 35.0 & 20.0 \\
\midrule
2Wiki & No rationale & 78.0 & 93.0 & 3.0 & 4.0 & 50.0 \\
2Wiki & Faithful & 77.0 & 93.0 & -- & -- & -- \\
2Wiki & Harmless & 77.0 & 93.0 & 0.0 & 2.0 & 0.0 \\
2Wiki & Entity swap & 66.0 & 42.0 & 15.0 & 55.0 & 20.0 \\
2Wiki & Answer conflict & 60.0 & 48.0 & 30.0 & 51.0 & 35.3 \\
\bottomrule
\end{tabular}
\caption{Main diagnostic results. Values are percentages. $\Delta$Ans and $\Delta$Supp are paired sensitivities relative to the faithful-rationale condition; $P(\Delta A\mid\Delta S)$ asks how often answer changes accompany support flips. F1 follows the same qualitative pattern and is omitted for space.}
\label{tab:main-results}
\end{table*}

\subsection{Harmless and No-Rationale Baselines}

Table~\ref{tab:main-results} makes the two controls explicit. First, harmless paraphrases behave like faithful rationales: support changes in only 4\% of MuSiQue examples, 1\% of HotpotQA examples, and 2\% of 2Wiki examples, with no significant paired differences (all McNemar $p\geq0.727$). HotpotQA's faithful support rate is already at 100\%, so the 1\% harmless change there is mainly a non-degradation check rather than evidence about additional headroom; MuSiQue and 2Wiki provide the more informative harmless-control comparisons. Together with the paraphrase audit and test--retest floor in Section~5.1, these controls rule out generic wording brittleness. Second, the no-rationale condition is close to the faithful condition on all three datasets: EM changes by at most one point, and support rates differ by at most three points. Thus, once evidence and a candidate answer are fixed, the faithful rationale is not a strong positive answer-selection signal in this verifier role. Its larger measurable effect appears when the communicated rationale contains factual errors.

\subsection{Blind Prompt: The Effect Is Not Mere Prompt Compliance}

\begin{table}[t]
\centering
\scriptsize
\setlength{\tabcolsep}{3.0pt}
\begin{tabular}{@{}lrrr@{}}
\toprule
Blind-prompt dataset & Harmless & Entity & Conflict \\
\midrule
MuSiQue-200 & 2.5 & 17.5$^{***}$ & 22.0$^{***}$ \\
HotpotQA-100 & 0.0 & 10.0$^{**}$ & 21.0$^{***}$ \\
2Wiki-100 & 0.0 & 17.0$^{***}$ & 14.0$^\dagger$ \\
\bottomrule
\end{tabular}
\caption{Support-change rates under a blind verifier prompt that omits explicit rationale-faithfulness instructions. Values are percentages relative to the faithful-rationale condition. $^\dagger p<0.06$, $^{**}p<0.01$, $^{***}p<0.001$ by paired McNemar tests.}
\label{tab:blind-prompt}
\end{table}

The blind-prompt ablation addresses the strongest prompt-confound concern. This prompt removes the explicit instruction to evaluate rationale faithfulness and asks only whether the answer is supported. As Table~\ref{tab:blind-prompt} shows, harmless paraphrases again leave support nearly unchanged (0--2.5\%), while corrupted rationales still change support in 10--22\% of examples. The effect is weaker than under the rationale-checking prompt in Table~\ref{tab:main-results}, so prompt design changes the size of the effect. However, the three-dataset blind-prompt results show that the effect is not merely created by an explicit faithfulness-checking instruction.

\subsection{Semantic Corruption Changes Support Judgments}

Corrupted rationale conditions produce a sharply different pattern. Relative to faithful rationales, entity-swapped corruptions change support judgments in 34--55\% of examples, and answer-conflicting corruptions change support in 35--51\%. All corrupted-condition comparisons are significant ($p<10^{-9}$ on every dataset). Support rates show the same separation, dropping from 85.5--100.0\% under faithful rationales to 42.0--66.0\% under corrupted rationales. The verifier notes also point to factual conflict rather than surface instability: among corrupted cases that flip support, 65--79\% of notes mention the corrupted target value or entity, and 94--99\% explicitly mention a rationale conflict, inconsistency, or incorrect statement. Together, Table~\ref{tab:main-results} and Figure~\ref{fig:support-answer} support the main diagnostic conclusion: verifier behavior is not broadly brittle to rewording, but it is sensitive to factual errors in the rationale field.

\subsection{Final Answers Stay More Stable Than Support}

Across all five conditions, final answers remain the same in 85.5\% of MuSiQue examples, 91.0\% of HotpotQA examples, and 67.0\% of 2Wiki examples. By contrast, support judgments remain the same in only 40.5\%, 50.0\%, and 31.0\% of examples. The coupling column in Table~\ref{tab:main-results} gives a normalized view that does not depend on subtracting two rates with different output spaces: under corrupted rationales, only 2.9--35.3\% of support flips are accompanied by answer changes, so 64.7--97.1\% of such support flips occur while the answer remains fixed.

This comparison has a structural asymmetry: the verifier is anchored to a candidate answer, while support is a binary judgment it must recompute. We therefore do not rely on a raw difference between answer-span changes and support-label changes. To stress-test the anchor, we run an answer-free HotpotQA-100 arm in which the model receives evidence and rationale but no candidate answer. Removing the anchor increases answer movement, but corrupted rationales still move support more than answers: 60\% versus 13\% for entity swaps, and 50\% versus 22\% for answer conflicts. Harmless paraphrases remain stable (1\% support change, 3\% answer change). Thus the main pattern is not solely an artifact of copying a supplied candidate answer, although the diagnostic remains a post-evidence setting rather than a full end-to-end generation study. Our primary evidence comes from the pattern across controls, the conditional coupling rates, the answer-free stress test, and boundary cases. 2Wiki answer-conflicting corruptions change final answers in 30\% of examples, corresponding to the EM drop from 77.0 to 60.0 in Table~\ref{tab:main-results}, and SciFact corruptions change final labels in 34--45\% of examples. The answer channel can move when the task structure makes it easy to move; support still moves more strongly and more consistently in the QA verifier setting.

As an additional check, $P(\Delta\text{answer}\mid\neg\Delta\text{support})$ ranges from 1.5\% to 24.5\% across datasets and corruption types. Answer changes are somewhat more likely when support flips, especially for answer-conflicting rationales, but support flips often occur without answer changes. This supports the channel-level reading: support assessment and answer selection are coupled but not identical outputs.

\subsection{The Support Bit Combines Answer and Rationale Checks}

The split-channel check asks what information the single support bit combines. It does not remove every candidate-answer anchoring concern, since final answers are still provided to the verifier; instead, it makes answer support and rationale faithfulness separate binary judgments. On HotpotQA-100, we replace the single \texttt{supported} field with \texttt{answer\_supported}, \texttt{rationale\_faithful}, and \texttt{overall\_supported}. Faithful and harmless rationales remain stable at 100\% across all three channels. Under corrupted rationales, the channels separate sharply: answer support remains high (90\% for entity swaps; 84\% for answer conflicts), while rationale faithfulness falls to 12\% and 11\%. Overall support follows rationale faithfulness almost exactly (12\% and 9\%), not answer support. This suggests that a single \texttt{supported} bit is not simply an answer-support label; in verifier pipelines that check rationales, it can act like a joint check over answer support and rationale faithfulness. By itself, this split-prompt pattern could reflect the explicit split requested by the prompt; the link back to the original single-channel verifier comes from the cases below.

Among the original single-channel correct-answer penalties on HotpotQA-100, there are 24 entity-swap and 19 answer-conflict cases where the original verifier rejected support while the answer remained correct. The split verifier still marks the answer itself supported in 16 and 10 cases, respectively, while rejecting the overall bundle. This supports the extrapolation that many original unsupported decisions were driven by the rationale channel rather than answer support. Conversely, among single-channel still-supported cases, it marks the rationale unfaithful in 54/66 and 55/65 cases. These broader still-supported sets are not the same as the audited corruption-overtrust category in Section~5.1; they are the full HotpotQA-100 cases where a corrupted rationale did not make the original verifier reject support. Splitting the output distinguishes ``answer is supported but rationale is bad'' from ``the whole output is acceptable.''

\subsection{Why the Dissociation Happens}

The dissociation arises because the final-answer channel and support channel use rationales differently. The answer channel can often be resolved locally from the fixed evidence and candidate answer: if the answer span is directly recoverable, a corrupted rationale need not change the string returned by the verifier. The support channel instead asks whether the full output is consistent with the evidence. Once the rationale is part of what is being verified, a contradiction inside that rationale can reasonably make an otherwise correct answer unsupported as a system output.

This does not mean that every support change is a failure. If the system output includes a corrupted rationale, rejecting the bundle can be the correct behavior: the answer may be right, but the communicated explanation is not a supported claim. The diagnostic is useful because it separates appropriate rejection from two auditable miscalibrations: \emph{correct-answer penalty}, where a correct answer is penalized by a bad rationale when the desired output is just the answer, and \emph{corruption overtrust}, where the verifier keeps accepting an invalid rationale. We quantify these modes through human audit in Section~5.1. We also observe \emph{local answer dominance} qualitatively: the answer is so directly supported that the verifier discounts an upstream rationale corruption. These patterns would be invisible under final EM alone, which is why the diagnostic reports answer and support channels separately.

\subsection{Depth and Task Structure Bound the Claim}

Rationale sensitivity grows with reasoning depth in MuSiQue. Entity-swapped rationales change support for 37.5\% of 2-hop, 42.9\% of 3-hop, and 48.5\% of 4-hop questions; answer-conflicting rationales change support for 31.7\%, 54.0\%, and 54.5\%, respectively. This suggests that the rationale message becomes more verification-relevant when evidence chains are longer. The natural-rationale audit points in the same direction: 8/11 unfaithful unperturbed rationales come from MuSiQue.

SciFact clarifies a task boundary. Harmless paraphrases remain stable, but corrupted rationales change the final label more often: 34\% for entity-swapped and 45\% for answer-conflicting rationales. This is not a contradiction; it is what the framework predicts when the task label is itself a support judgment. In extractive QA, answer selection and support assessment can separate because the answer span may remain recoverable from evidence even when the rationale is invalid. 2Wiki is an intermediate case: many comparison-style questions have a small answer space, so answer-conflicting rationales can directly flip the final span more often than on MuSiQue or HotpotQA. In fact verification, the final output is already a support/refute judgment, so rationale sensitivity naturally propagates into the task label. Thus QA systems should report answer stability and support stability separately, while claim-verification systems should treat label stability as the task-level form of support sensitivity.

\paragraph{Design implication.}
If rationales mainly changed answer selection, the natural response to an unsupported output would be answer regeneration. Our results suggest a different design rule for agent communication interfaces: before relying on a message field, test what receiver behavior it actually changes. After an answer is formed, rationale messages should be treated as claims to verify against evidence. A deployment should first test whether the rationale channel is active, harmless, or ignored. If it is active, systems should distinguish rationale repair, evidence rechecking, support escalation, and answer regeneration rather than collapsing them into one binary failure state. If it is ignored, as in the DeepSeek-R1 boundary case, passing rationales to later roles may add interface complexity without communication value.

\section{Validation and Robustness}

\begin{figure*}[t]
\centering
\includegraphics[width=0.75\textwidth]{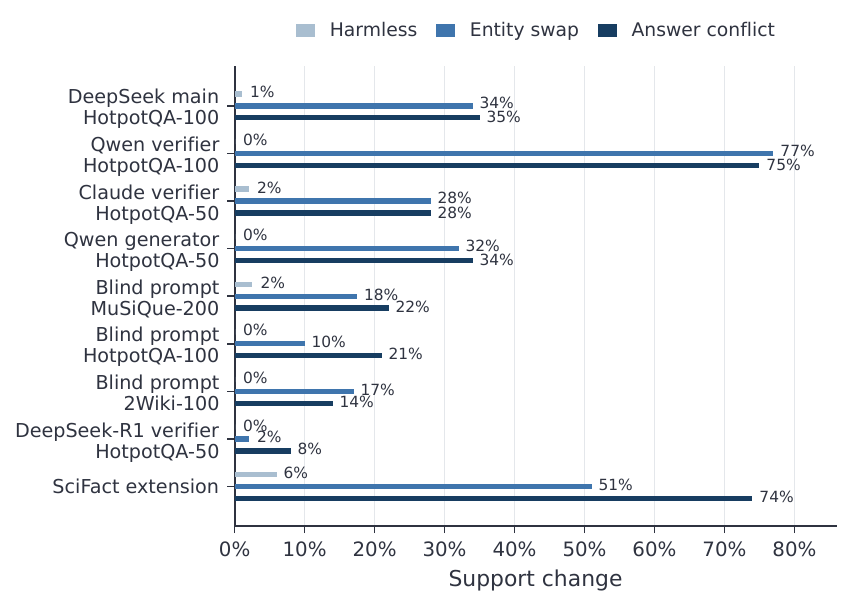}
\caption{Robustness and boundary checks. Harmless paraphrases remain stable; corrupted-rationale responsiveness varies across verifier, generator, prompt, and task-domain checks.}
\label{fig:robustness}
\end{figure*}

\subsection{Audits and Cross-Model Checks}

Human audits support the proposed mechanism. In the corrupted-rationale audit, 13/42 valid corruptions are correct-answer penalty cases: the answer remains correct but the verifier marks it unsupported because the rationale is corrupted. Another 16/42 are corruption-overtrust cases: the verifier keeps marking the output supported despite a valid corruption. The remaining 13/42 are not failures of the diagnostic; they are cases where the corruption also changes answer correctness or where rejection is an appropriate response rather than evidence of answer/support dissociation. A separate blind 50-item human support audit confirms that the two failure modes are not merely model-label noise. Among 20 corrupted examples where the model flips support to unsupported, the blind label is also unsupported in 19 cases. In the opposite direction, among 10 corrupted examples where the model still predicts supported, the blind label is unsupported or unclear in 9 cases, indicating that corruption overtrust is a real verifier failure mode rather than an artifact of model labels.

The harmless-paraphrase audit shows the complementary control. On the full audited sample, harmless support change is 2\%; after filtering to acceptable paraphrases it falls to 0\%, while corrupted changes remain 39.5\% for entity swaps and 34.9\% for answer conflicts. In the strict-valid subset, harmless support change is also 0\%, while both corrupted conditions change support in 36.1\% of examples.

A same-input test--retest run gives a low noise floor. We rerun the faithful-rationale condition once for all 400 QA examples using the same model, prompt, inputs, and temperature. The support bit changes in 6/400 examples (1.5\%; 3.0\% on MuSiQue, 0\% on HotpotQA and 2Wiki), final-answer strings change in 3/400 (0.75\%), and EM changes in 0/400. Thus the 34--55\% corrupted-rationale support changes are far above ordinary API-level output instability, while harmless changes remain at or near the retest floor.

Figure~\ref{fig:robustness} summarizes robustness checks, each aimed at a different alternative explanation. To test verifier-family dependence, we replace the DeepSeek verifier with Qwen and Claude. Qwen shows an even stronger separation than the main verifier: harmless support change is 0\%, while corrupted-rationale support change is 75--77\%. Claude shows the same direction with smaller effect sizes: harmless change is 2\%, while each corrupted condition changes support in 28\% of examples. To test generator dependence, we use Qwen as the generator/reasoner and DeepSeek as verifier; harmless change is again 0\%, while corrupted changes are 32--34\%.

\subsection{Reasoning Verifiers Can Make the Channel Inert}

DeepSeek-R1 is the main boundary case. It shows 96\% answer stability and no harmless support changes, but corrupted rationales change support in only 2--8\% of examples. This does not invalidate the diagnostic; it reveals a different failure mode, which we call channel inertia. A reasoning verifier may reconstruct support from evidence more independently and discount the externally supplied rationale field. That makes it robust to corrupted messages, but it also means the rationale channel between roles has little force. For system design, this is not simply a stronger-verifier success case: it shows that as verifiers become more self-sufficient, explicit rationale communication may become inert unless the verifier is designed to inspect the message.

\section{Related Work}

Faithfulness work distinguishes rationalizations from explanations that reflect the process that produced a decision \citep{jacovi2020towards}. LLM rationales can be plausible but unfaithful \citep{turpin2023language,cornish2025examining}, and counterfactual or perturbation tests ask whether explanations are faithful to a model's own prediction behavior \citep{atanasova2023faithfulness,lanham2023measuring,tutek2025measuring,selfauditing2026,cotreuse2026}. Other work cautions that many such tests measure output-level self-consistency rather than hidden reasoning \citep{parcalabescu2024measuring}. We study a different object: a rationale after it crosses a role boundary. The unit is the message between roles, and the measured outputs are separated into answer selection and support assessment.

Verifier-based reasoning uses verifiers, self-consistency, and process supervision to select or check solutions \citep{cobbe2021training,wang2023selfconsistency,lightman2024lets}. LLM-as-judge and QA reevaluation work shows sensitivity to prompts, references, answer forms, and rubrics \citep{zheng2023judging,referenceguided2024,clev2025,reassessing2025extractive}; RAG evaluation separates faithfulness, answer relevance, and context use \citep{lewis2020rag,es2024ragas}.

Multi-agent communication work studies what information should pass between agents and how bad messages propagate. Recent work characterizes category-level information exchange \citep{chun2026what}, error cascades \citep{xie2026spark,sakib2026preventing}, and harmful agreement or sycophancy in debate-style systems \citep{wynn2025talk,pitre2025consensagent,hao2026flips}. These studies analyze coordination across agents or debate rounds. We focus on a narrower interface: a single rationale field crossing a reasoner--verifier boundary, where message perturbations reveal whether the receiver treats the field as answer evidence, support evidence, or inert text.

\section{Limitations}

The diagnostic is designed for interface analysis: it holds evidence and candidate answer fixed so that changes can be attributed to the rationale message. This fixed-context design is the source of its causal clarity. End-to-end evaluations remain complementary when the research question concerns how retrieval, reasoning, answer generation, and verification change together.

The paired intervention design requires multiple verifier calls and perturbation variants per example. We therefore prioritize controlled comparisons and statistical rigor over benchmark scale: the study uses 400 primary examples across three datasets, plus cross-model checks and audits, and relies on paired tests, a repeated-run noise floor, and targeted human audits rather than treating the result as a large benchmark.

Perturbations are generated by LLMs and audited on samples rather than exhaustively labeled. The corrupted-rationale and harmless-paraphrase audits support intervention quality, and the natural-rationale audit shows that rationale/evidence mismatches also occur in unperturbed pipeline outputs. Support labels are model-generated verifier judgments, which matches the paper's target: communication behavior in verifier pipelines. The 50-item human support audit is enriched for diagnostic cases and should not be read as a global verifier-accuracy estimate; its role is to test whether the central support changes align with human judgments. The primary human support labels come from a blind second annotator, with the earlier human pass used to report two-annotator agreement.

The primary experiments study multi-hop QA, with SciFact as a smaller task-domain extension. This makes the conclusions strongest for role-specialized QA and verification pipelines; tool-use, planning, and open-ended collaborative agents are natural targets for applying the same diagnostic.

The main verifier prompt asks the model to check rationales because it models systems that intentionally pass rationales for verification. A blind-prompt ablation shows a smaller but persistent message effect. This makes prompt design part of the phenomenon: explicit rationale checking amplifies the channel, while blind prompts reveal whether the channel is active without that instruction.

Our experiments are black-box behavioral tests. They show when a verifier changes its answer or support judgment under a controlled message intervention. Explaining the internal computation behind those changes is a separate mechanistic question, especially for the DeepSeek-R1 boundary case: the model may reconstruct support from evidence more independently, follow different prompt priors, or ignore external rationale fields.

A verifier's binary support label is a decision made by a later verifier under a specific prompt and evidence context, not a formal proof label. This is the intended object of study: communication behavior in role-specialized LLM pipelines. Systems requiring proof-level guarantees can combine this diagnostic with symbolic checks, retrieval audits, or human review.

Most diagnostic conditions assume that the candidate answer is already formed before verification. This matches common verifier pipelines and lets the protocol isolate the reasoner-to-verifier message. The HotpotQA-100 answer-free stress test probes how the pattern changes when the answer anchor is removed. The split-channel check is likewise limited to HotpotQA-100; it is a mechanism check for the single-support-bit interpretation and a template for future multi-dataset channel probes.

\bibliography{references}

\appendix

\section{Audit Details}

\subsection{Human Support Audit}

The 50-item human support audit is enriched for diagnostic cases rather than sampled to estimate global verifier accuracy. It contains 20 corrupted examples where the model flips support to unsupported, 10 corrupted examples where the model still marks the output supported, 10 harmless controls, and 10 faithful baselines. To avoid anchoring, a second annotator performs a blind pass that hides the model support label, model confidence, verifier note, sampling bucket, and condition name. The blind annotator sees only the question, evidence passages, candidate/final answer, and displayed rationale, then labels the output as supported, unsupported, or unclear under that rationale. We use the blind pass as the primary human label. Blind model-human agreement is 74\%, with Cohen's $\kappa=0.52$ over three labels; the lower model-human agreement is concentrated in the corrupted-supported bucket, where the model accepts corrupted rationales that the blind annotator rejects. The two human annotation passes agree on 88\% of items ($\kappa=0.77$), or 93.6\% with $\kappa=0.87$ after excluding unclear labels.

\begin{center}
\centering
\small
\resizebox{0.94\columnwidth}{!}{%
\begin{tabular}{@{}lrrrr@{}}
\toprule
Bucket & $n$ & Agreement & Human supp. & Model supp. \\
\midrule
Support flip & 20 & 95\% & 5\% & 0\% \\
Corrupted supported & 10 & 10\% & 10\% & 100\% \\
Faithful baseline & 10 & 90\% & 90\% & 100\% \\
Harmless control & 10 & 80\% & 80\% & 90\% \\
\bottomrule
\end{tabular}
}
\captionof{table}{Blind human support audit by enriched sampling bucket.}
\end{center}

\subsection{Harmless-Paraphrase Filtering}

The harmless-paraphrase audit labels each paraphrase as strict-valid, minor-change, or invalid. The main text reports both full-sample and filtered robustness checks. Filtering strengthens the harmless control: original-to-harmless support change falls to 0\% in both acceptable and strict-valid subsets, while corrupted-rationale conditions remain disruptive.

\begin{center}
\centering
\small
\resizebox{0.94\columnwidth}{!}{%
\begin{tabular}{@{}lrrrr@{}}
\toprule
Subset & $n$ & Harmless & Entity swap & Answer conflict \\
\midrule
All audited & 50 & 2.0\% & 42.0\% & 38.0\% \\
Valid + minor & 43 & 0.0\% & 39.5\% & 34.9\% \\
Strict valid & 36 & 0.0\% & 36.1\% & 36.1\% \\
\bottomrule
\end{tabular}
}
\captionof{table}{Audit-filtered support-change rates relative to the original rationale condition.}
\end{center}

\subsection{Natural Rationale Inconsistency Audit}

To test ecological validity, we also audit 100 unperturbed reasoner-generated rationales sampled across MuSiQue, HotpotQA, and 2Wiki. This audit asks whether the original pipeline naturally produces rationale/evidence or rationale/answer inconsistencies, rather than only under our synthetic corruptions. Table~\ref{tab:natural-audit} shows that 66/100 examples are consistent, while 11/100 contain unfaithful rationale errors. The remaining non-none categories mostly identify evaluation artifacts or correct rejections: 16/100 are metric false negatives, 5/100 are correct rejections of wrong answers, 1/100 is a verifier false negative, and 1/100 reflects a question/gold flaw. Thus, the intervention studies a naturally occurring failure mode, but the audit also prevents over-attributing ordinary EM errors to rationale unfaithfulness.

\begin{center}
\centering
\small
\resizebox{0.94\columnwidth}{!}{%
\begin{tabular}{@{}lrrrr@{}}
\toprule
Category & All & MuSiQue & HotpotQA & 2Wiki \\
\midrule
Consistent & 66 & 16 & 23 & 27 \\
Metric false negative & 16 & 4 & 8 & 4 \\
Unfaithful rationale & 11 & 8 & 2 & 1 \\
Correct rejection & 5 & 4 & 0 & 1 \\
Verifier false negative & 1 & 1 & 0 & 0 \\
Question/gold flaw & 1 & 1 & 0 & 0 \\
\bottomrule
\end{tabular}
}
\captionof{table}{Natural rationale inconsistency audit over 100 unperturbed reasoner outputs. Categories are mutually exclusive error-type labels.}
\label{tab:natural-audit}
\end{center}

\section{Qualitative Case Details}

Table~\ref{tab:cases} summarizes three cases. Case 001 remains answer-correct but is rejected because the rationale moves Fiorello La Guardia from New York to Chicago. Case 011 has a corrupted headquarters link, but the answer remains recoverable, so the verifier keeps support. Case 018 has a locally supported answer span, so the verifier does not penalize a corrupted upstream military-branch relation. These cases illustrate why answer accuracy alone cannot tell whether a rationale message is being used for support assessment.

\begin{center}
\centering
\footnotesize
\setlength{\tabcolsep}{2.4pt}
\begin{tabular}{@{}>{\raggedright\arraybackslash}p{0.13\columnwidth}
                >{\raggedright\arraybackslash}p{0.30\columnwidth}
                >{\raggedright\arraybackslash}p{0.23\columnwidth}
                >{\raggedright\arraybackslash}p{0.24\columnwidth}@{}}
\toprule
Case & Perturbation & Outcome & Pattern \\
\midrule
001 & Mayoral city: New York to Chicago & Correct answer rejected & Correct-answer penalty \\
011 & UMG HQ: Santa Monica to New York City & Correct answer accepted & Corruption overtrust \\
018 & Military branch: Navy to Army & Direct answer still accepted & Local answer dominance \\
\bottomrule
\end{tabular}
\captionof{table}{Qualitative audit examples illustrating two auditable failure modes and one qualitative dissociation pattern.}
\label{tab:cases}
\end{center}

\section{Answer-Free Stress Test}

Table~\ref{tab:answer-free} reports the HotpotQA-100 stress test that removes the candidate answer from the verifier input. The model receives only the question, evidence, and optional rationale, then generates its own answer and support judgment. This test directly probes whether answer stability in the main diagnostic is merely caused by copying a supplied candidate answer. Removing the anchor increases answer sensitivity, especially for answer-conflicting rationales, but support sensitivity remains larger than answer sensitivity under both corrupted conditions.

\begin{center}
\centering
\scriptsize
\setlength{\tabcolsep}{2.2pt}
\resizebox{0.98\columnwidth}{!}{%
\begin{tabular}{@{}lrrrrr@{}}
\toprule
Condition & EM & Support & $\Delta$Ans & $\Delta$Supp & $P(\Delta A\mid\Delta S)$ \\
\midrule
No rationale & 64.0 & 93.0 & 14.0 & 4.0 & 50.0 \\
Faithful & 68.0 & 97.0 & -- & -- & -- \\
Harmless & 66.0 & 98.0 & 3.0 & 1.0 & 0.0 \\
Entity swap & 63.0 & 39.0 & 13.0 & 60.0 & 13.3 \\
Answer conflict & 60.0 & 47.0 & 22.0 & 50.0 & 26.0 \\
\bottomrule
\end{tabular}
}
\captionof{table}{Answer-free verifier stress test on HotpotQA-100. Values are percentages. The verifier receives no candidate answer and must answer from evidence and optional rationale.}
\label{tab:answer-free}
\end{center}

\section{Split-Channel Verification Details}

Table~\ref{tab:split-channel} reports the HotpotQA-100 split-channel verifier probe. The intervention uses the same five rationale conditions as the main experiment, but asks the verifier to separately output answer support, rationale faithfulness, and overall support. The key pattern is that corrupted rationales mostly affect the rationale-faithfulness and overall channels, while answer support remains substantially higher.

\begin{center}
\centering
\scriptsize
\setlength{\tabcolsep}{2.0pt}
\resizebox{0.98\columnwidth}{!}{%
\begin{tabular}{@{}lrrrr@{}}
\toprule
Condition & EM & Answer supp. & Rationale faithful & Overall supp. \\
\midrule
No rationale & 68.0 & 97.0 & -- & 97.0 \\
Faithful & 68.0 & 100.0 & 100.0 & 100.0 \\
Harmless & 68.0 & 100.0 & 100.0 & 100.0 \\
Entity swap & 68.0 & 90.0 & 12.0 & 12.0 \\
Answer conflict & 66.0 & 84.0 & 11.0 & 9.0 \\
\bottomrule
\end{tabular}
}
\captionof{table}{Split-channel verifier probe on HotpotQA-100. Values are percentages.}
\label{tab:split-channel}
\end{center}

\clearpage
\section{Additional Visualizations}

Figure~\ref{fig:app-cross-task} shows the cross-task view of support change. We keep it in the appendix because the multi-hop QA values duplicate Figure~\ref{fig:support-answer}, while the SciFact boundary is also summarized in Figure~\ref{fig:robustness}.

\begin{center}
\centering
\includegraphics[width=\columnwidth]{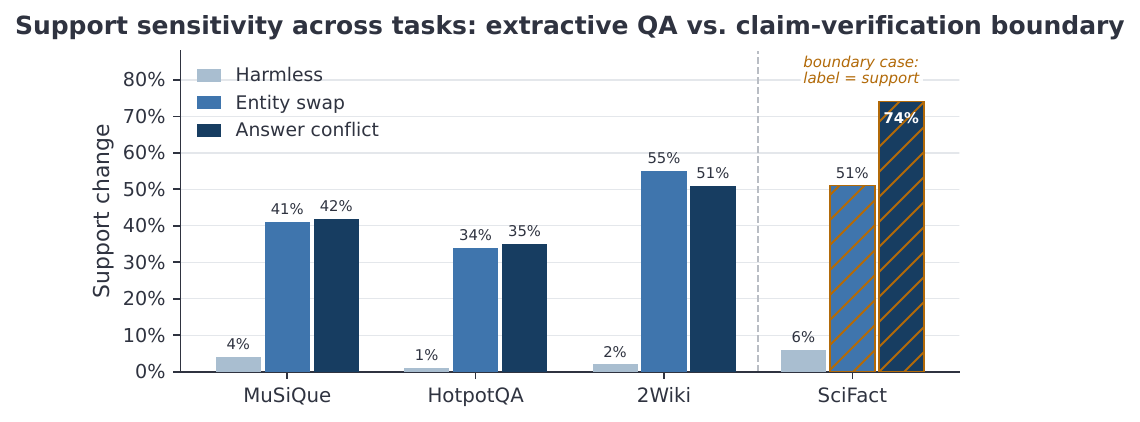}
\captionof{figure}{Support change across QA and claim verification. Multi-hop QA shows strong support sensitivity with relatively stable final answers, while SciFact is a boundary case where support sensitivity propagates into the final label because the task output is itself a verification label.}
\label{fig:app-cross-task}
\end{center}

Figure~\ref{fig:app-supported-rate} shows the raw verifier supported rate under each rationale condition. It complements Figure~\ref{fig:support-answer} by showing that harmless paraphrases closely track faithful rationales, while both corruption types reduce support rates.

\begin{center}
\centering
\includegraphics[width=\columnwidth]{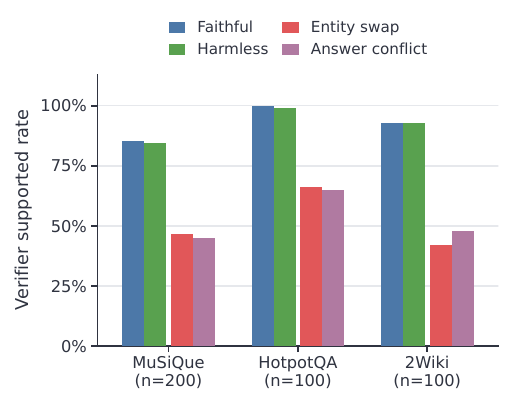}
\captionof{figure}{Raw verifier supported rate by rationale condition. Faithful and harmless conditions remain close, while corrupted rationale conditions substantially reduce support.}
\label{fig:app-supported-rate}
\end{center}

Figure~\ref{fig:app-stability} reports the fraction of examples whose final answer and support judgment remain unchanged across all five rationale conditions. The figure makes the answer/support dissociation visually explicit: answers are much more stable than support judgments.

\begin{center}
\centering
\includegraphics[width=\columnwidth]{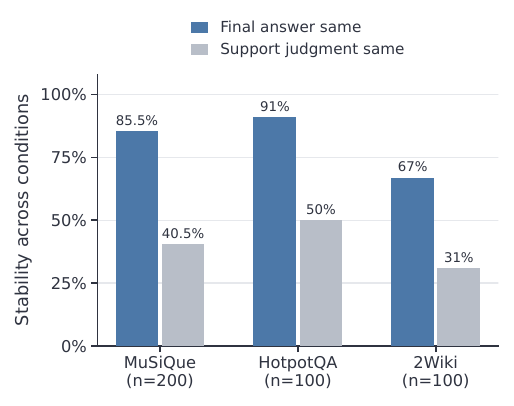}
\captionof{figure}{Answer and support stability across all five rationale conditions. Final answers are generally more stable than support judgments, with the strongest support instability on MuSiQue and HotpotQA.}
\label{fig:app-stability}
\end{center}

Figure~\ref{fig:app-audit-filtered} visualizes the harmless-paraphrase audit. After filtering to valid or strict-valid paraphrases, harmless changes disappear while corrupted-rationale changes remain large.

\begin{center}
\centering
\includegraphics[width=\columnwidth]{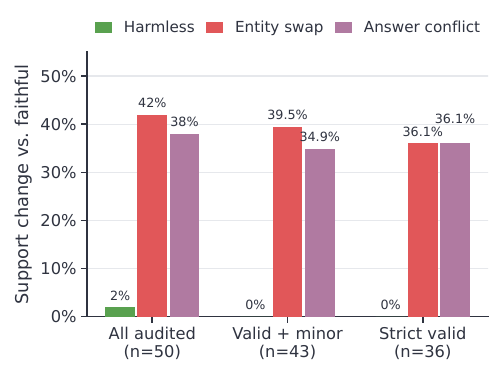}
\captionof{figure}{Audit-filtered harmless-control robustness. Harmless paraphrases induce no support changes after filtering, whereas corrupted rationales remain disruptive.}
\label{fig:app-audit-filtered}
\end{center}

\section{Model and Decoding Details}
\label{sec:model-details}

Table~\ref{tab:model-details} lists the hosted model identifiers used in the experiments. All calls use temperature 0. Standard verifier and generator calls cap output at 512 tokens; the reasoning-verifier check uses a larger cap because the hosted reasoning endpoint may emit longer structured responses. Provider-side parameter counts and exact weight snapshots are not publicly disclosed for these API models, so we report the exact API model strings and the execution window instead.

\begin{center}
\centering
\footnotesize
\setlength{\tabcolsep}{3.2pt}
\begin{tabular}{@{}>{\raggedright\arraybackslash}p{0.28\columnwidth}
                >{\raggedright\arraybackslash}p{0.64\columnwidth}@{}}
\toprule
Label in paper & Hosted API identifier and settings \\
\midrule
DeepSeek main & \texttt{deepseek-v4-flash}; generator/verifier; Jun.--Jul. 2026; temperature 0; max output 512. \\
Qwen verifier/generator & \texttt{qwen3.7-plus-2026-05-26}; robustness checks; Jul. 2026; temperature 0; max output 512. \\
Claude verifier & \texttt{claude-haiku-4-5}; recheck/robustness; Jul. 2026; temperature 0; max output 512. \\
DeepSeek-R1 verifier & \texttt{deepseek-reasoner}; reasoning boundary check; Jul. 2026; temperature 0; max output 4096. \\
\bottomrule
\end{tabular}
\captionof{table}{Hosted model identifiers and decoding settings. Temperature is 0 for all rows. Parameter counts and exact provider snapshots are not disclosed by the hosted APIs.}
\label{tab:model-details}
\end{center}

\section{Verifier Prompt and Output Schema}

All diagnostic verifier calls use the same evidence passages and candidate answer for a given example; only the rationale field changes across conditions. The verifier returns a structured record with four fields: final answer, binary support label, confidence, and a short verification note. In the no-rationale condition, the rationale field is omitted. In the faithful, harmless, entity-swapped, and answer-conflicting conditions, the verifier receives the corresponding rationale text in the same slot.

The main verifier prompt asks whether the candidate answer is supported by the evidence and whether the displayed rationale is consistent with that evidence. The blind-prompt ablation removes the explicit instruction to evaluate rationale faithfulness and asks only for answer support. This ablation is used to test whether rationale sensitivity is merely induced by verifier wording; corrupted rationales still change support judgments, but with smaller effects.

The perturbation protocol is therefore a controlled message intervention rather than a re-generation experiment: retrieval evidence, candidate answer, verifier model, decoding setting, and output schema are fixed while the rationale message is varied.

\end{document}